\documentclass[sigconf,prologue,table]{acmart}

\usepackage{amsmath} 

\usepackage{graphicx} 
\usepackage{multirow}
\usepackage{arydshln}
\hypersetup{
    backref=page,
    colorlinks=true,
    linkcolor=blue,
    citecolor=blue
}
\AtBeginDocument{%
  }

\setcopyright{acmlicensed}
\copyrightyear{2025}
\acmYear{2025}
\acmDOI{XXXXXXX.XXXXXXX}
\acmConference[SLTAT '25]{}{Sept 17--19,
  2025}{Berlin, DE}
\acmISBN{978-1-4503-XXXX-X/2018/06}

\begin{document}

\title{Hierarchical Feature Alignment for Gloss-Free Sign Language Translation}

\author{Sobhan Asasi}
\email{s.asasi@surrey.ac.uk}
\affiliation{%
  \institution{University of Surrey}
  \city{Guildford}
  \country{United Kingdom}
}

\author{Mohamed Ilyes Lakhal}
\email{m.lakhal@surrey.ac.uk}
\affiliation{%
  \institution{University of Surrey}
  \city{Guildford}
  \country{United Kingdom}
}

\author{Richard Bowden}
\email{r.bowden@surrey.ac.uk}
\affiliation{%
  \institution{University of Surrey}
  \city{Guildford}
  \country{United Kingdom}
}

\renewcommand{\shortauthors}{Trovato et al.}

\begin{abstract}
Sign Language Translation (SLT) attempts to convert sign language videos into spoken sentences. However, many existing methods struggle with the disparity between visual and textual representations during end-to-end learning. Gloss-based approaches help to bridge this gap by leveraging structured linguistic information. While, gloss-free methods offer greater flexibility and remove the burden of annotation, they require effective alignment strategies. Recent advances in Large Language Models (LLMs) have enabled gloss-free SLT by generating text-like representations from sign videos. In this work, we introduce a novel hierarchical pre-training strategy inspired by the structure of sign language, incorporating pseudo-glosses and contrastive video-language alignment. Our method hierarchically extracts features at frame, segment, and video levels, aligning them with pseudo-glosses and the spoken sentence to enhance translation quality. Experiments demonstrate that our approach improves BLEU-4 and ROUGE scores while maintaining efficiency.
\end{abstract}

\ccsdesc[500]{Computing methodologies~Computer vision}

\keywords{Sign Language Translation, Vision-Language Pre-training, Large Language Models}
\maketitle

\section{Introduction}
Sign language uses hand gestures, body movements, and facial expressions for communication \cite{p44, p47}. \mbox{Sign language translation (SLT)} aims to convert these visual elements into spoken sentences, enabling communication with non-signers. However, SLT faces challenges in accurately recognizing the components of sign and bridging the gap between visual and textual representations \cite{p16, csgcr}.

SLT approaches typically fall into two categories: \mbox{gloss-based} and \mbox{gloss-free}. Gloss-based methods rely on intermediate representations, where sign videos are first translated into a sequence of glosses, written representations of signs using corresponding words in a spoken language, before generating spoken sentences. In contrast, gloss-free methods directly map videos to text without an explicit gloss representation \cite{p6,p53}. While gloss-based approaches benefit from structured linguistic information, gloss-free methods bypass the need for gloss annotations, making them more flexible but often more challenging. The high cost and effort required for data collection and meticulous gloss annotation have led to increased interest in gloss-free SLT. Addressing this challenge is the primary focus of this paper.

The increasing use of Large Language Models (LLMs) has had a significant impact on this area. As a result, the latest gloss-free models utilize the remarkable language prediction capabilities of LLMs by converting sign language videos into text-like representations that LLMs can process effectively \cite{openai2024gpt4technicalreport,sign2gpt, gfslt-vlp}.
When LLMs are applied directly to visual features, they become mere transformers with pre-trained weights due to the large shift between the textual representations on which they were trained and the new visual inputs.
To bridge this gap, recent studies have introduced a pre-training phase before the SLT task, which aims to align visual and textual representations~\cite{sign2gpt, gfslt-vlp}.

Such pre-training strategies can take two forms: pseudo-gloss pretraining and contrastive visual-language pretraining. Pseudo-gloss pretraining generates intermediate text-like representations from sign videos to help LLMs process them more effectively. Contrastive visual-language pretraining, on the other hand, aligns visual and textual representations by learning a shared feature space, often using contrastive learning techniques to improve modality bridging.

\begin{figure*}[!t]
    \centering
    \includegraphics[width=\textwidth]{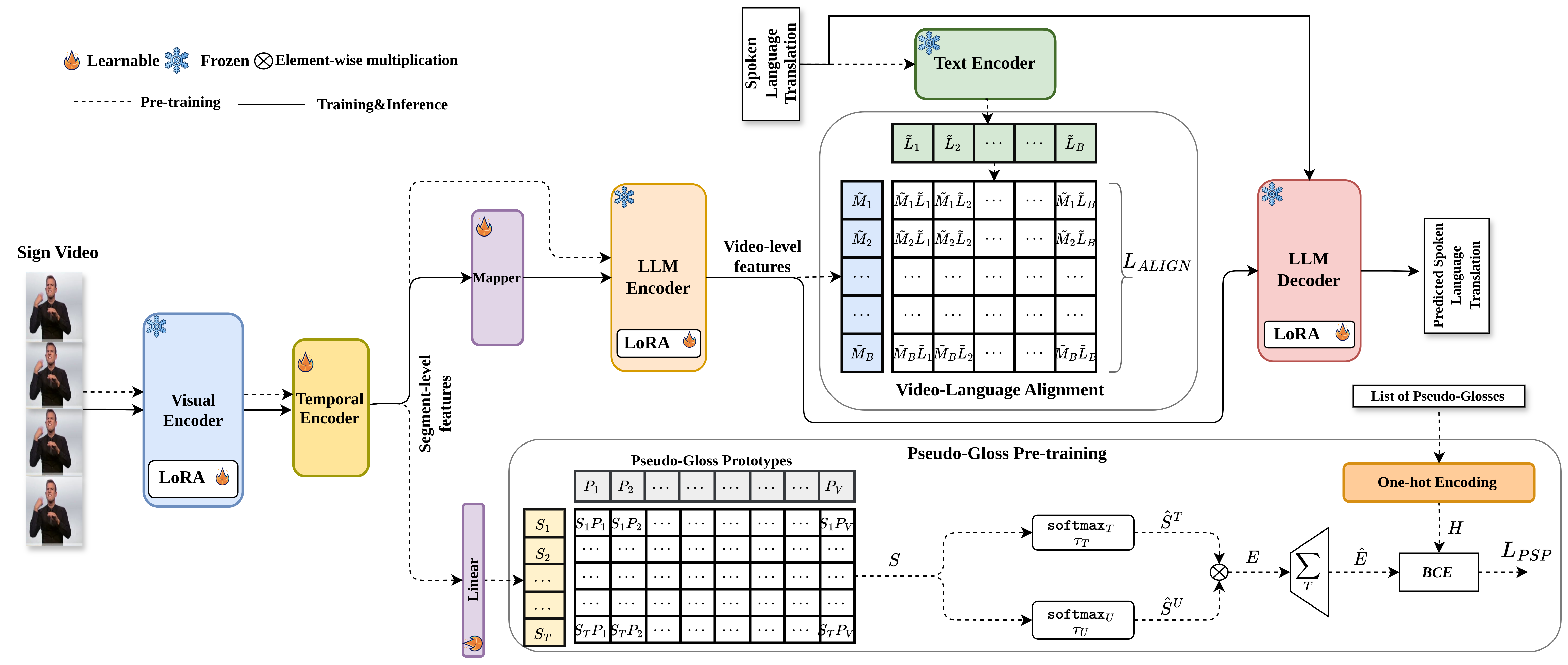} 
    \caption{\textbf{Model overview}. Our proposed pipeline we propose is divided into two phases: a comprehensive pre-training phase (dashed line in the figure) and a fine-tuning phase (solid line in the figure). During pre-training, the model bridges the gap between visual and textual representations at two levels. The pseudo-gloss pre-training matches visual features at the segment level with corresponding pseudo-glosses, enabling a fine-grained mapping between video segments and linguistic representations. At the same time, video-to-language alignment uses contrastive learning to map video-translation pairs at a global level, ensuring a coherent alignment between the video and target spoken sentence. During fine-tuning, segment-level features are processed by a mapper to improve their adaptation to the SLT task and boost overall performance.}
    \label{fig:example}
\end{figure*}

In this paper, we are inspired by the hierarchical structure of signs, in which nuanced hand and body movements form glosses, which in turn construct sentences.
As shown in Fig.~\ref{fig:example}, our method extracts features on three levels: frame, segment, and video. To bridge the modality gap and disparity between visual and textual representations, we apply two alignment strategies: \mbox{pseudo-gloss} pretraining, which aligns segment features with \mbox{pseudo-glosses}, and video-language alignment, which brings video features closer to the target spoken sentences. During the downstream SLT task, the pretraining alignments are removed, and the \mbox{video} features are directly fed into a decoder for final processing. Our proposed pipeline is more scalable than related approaches such as \mbox{GFSLT-VLP} \cite{gfslt-vlp}, as it removes the need for a decoder and avoids the need to update the text encoder weights during pretraining. The main contributions of this paper are as follows:
\begin{itemize}
    \item Our proposed model effectively bridges the gap between visual and textual representations on two different levels of sign video features.
    \item Our experiments demonstrate the effectiveness of our method, achieving improved gloss-free SLT performance on a benchmark dataset. Notably, our approach enhances both BLEU-4 and ROUGE scores, while preserving efficiency.
\end{itemize}
\section{Related Works}
\subsection{Gloss-based Sign Language Translation}
A dominant paradigm in Sign Language Translation (SLT) is the gloss-based approach, which first recognizes signs into glosses and then translates them to text~\cite{mmtlb, ts-slt, sltunet}. Recent works have pushed this forward. MMTLB~\cite{mmtlb} leverages large pre-trained language models for more fluent translation. TS-SLT~\cite{ts-slt} employs a dual-stream visual encoder (RGB and keypoints) for more robust feature extraction, and SLTUNET~\cite{sltunet} unifies related sub-tasks into a single model to improve learning efficiency and create more universal representations.
\subsection{Gloss-free Sign Language Translation}
 While gloss-based Sign Language Translation (SLT) methods are highly effective due to intermediate supervision from annotated glosses~\cite{sltunet, ts-slt}, they are limited by the costly and time-consuming nature of gloss creation, which is not feasible for all languages. To address this limitation, gloss-free SLT learns directly from video-text pairs. This approach is more scalable but contends with a larger modality gap, which typically results in lower performance.
To address this gap,
CSGCR~\cite{csgcr} predicts possible language words, generates
sentences based on these predictions, and selects the most appropriate sentence using cosine similarity with sign language images. Recently, GFSLT-VLP~\cite{gfslt-vlp} introduced a pre-training method that aligns sign language images with
spoken sentences, converting them into text-like features familiar to LLMs.
Leveraging this advancement, subsequent gloss-free models focus on generating more comprehensible features
for LLMs. FLa-LLM~\cite{fla-llm} introduces a two-step approach: initially training on visual information from sign language
images using a lightweight translation model and fine-tuning the LLM for SLT. Sign2GPT~\cite{sign2gpt} pre-trains a sign
encoder by aligning visual features with prototypes under the supervision of pseudo-glosses, words from spoken sentences via Parts-Of-Speech (POS) tagging, and then utilizes it for SLT.
\subsection{Weakly supervised gloss-free SLT}
While all gloss-free methods train directly on video-text pairs, weakly supervised approaches are distinct because they introduce an explicit, intermediate mechanism that uses the target sentence to create a proxy for alignment, rather than relying on a standard architecture to learn this implicitly. For example, TSPNet~\cite{tspnet} achieves this architecturally, using a temporal-semantic pyramid to structure the visual learning process. GASLT~\cite{gaslt} creates a soft alignment by using a 'Gloss Attention' mechanism, where the semantic content of the target sentence guides the model to focus on the most salient visual frames. More directly, VAP~\cite{vap} introduces a dedicated pre-training stage to explicitly align visual features with text embeddings from the parallel sentence, bridging the modality gap before the final translation task. These engineered mechanisms are the key feature that defines this weakly supervised category.
\subsection{Vision-Language Pre-training}
Visual-language pretraining (VLP) models are categorized into single-stream and dual-stream approaches. Single-stream models use a unified transformer encoder for both modalities, while dual-stream models process visual and text inputs separately, aligning them through contrastive learning or cross-attention. GFSLT-VLP \cite{gfslt-vlp} was the first to apply \mbox{CLIP-based} \mbox{dual-stream} pretraining to gloss-free SLT but lacks frame-level supervision. VAP \cite{vap} addresses this limitation by learning alignments between video frames and spoken words using weak supervision.


\section{Methodology}
In the following subsections, we present our model architecture (Fig. \ref{fig:example}) and describe the training strategy used for SLT.
\subsection{Architecture}
\textbf{Visual Encoder.} Our video-based sign language translation framework is built around an image encoder that extracts spatial features \( F \) from input video frames \mbox{\( X = \{x_0, x_1, \dots, x_{T^{*}}\} \)}, where \( T^{*} \) represents the total number of frames. These features, with a dimensionality of \( D^{*} \), are structured as \mbox{\( F \in \mathbb{R}^{T^{*} \times D^{*}} \)} and form the foundation for segment-level feature construction. We use DinoV2 \cite{dinov2learningrobustvisual}, a self-supervised Vision Transformer, as our image encoder. By fine-tuning the ViT-S/14 variant, we adapt it to sign language translation datasets, improving feature extraction and task performance. The class token's output serves as a feature vector, which is transformed linearly and batch-normalized to obtain \mbox{\( F \in \mathbb{R}^{T^{*} \times D^{*}} \)}, before being passed into the temporal encoder.

\textbf{Temporal Encoder.} We use frame features \( F \) as input to our temporal encoder to learn spatio-temporal sign representations. Since sign language sequences often contain hundreds of frames, we design a spatio-temporal transformer inspired by prior SLT approaches, incorporating two key modifications for efficiency and effectiveness. Similar to~\cite{sign2gpt}, we integrate local self-attention with a window size of seven, a proven technique for SLT \cite{gaslt}. Combined with downsampling, this extends the model’s temporal receptive field while reducing redundancy, ensuring compatibility with the LLM encoder. The resulting segment features are represented as \mbox{\( Z \in \mathbb{R}^{T \times D} \)}, where \( T \) is the downsampled sequence length and \( D \) is the new feature dimension. To effectively model both temporal and spatial information, we use Rotary Position Embedding (RoPE) \cite{rope} for positional encoding. RoPE encodes positional information into attention, preserving temporal and spatial relationships, enhancing the model's ability to capture spatio-temporal structure in sign language sequences.

\textbf{LLM Encoder.}
To the refine segment-level features into more expressive representations, we employ a LLM encoder, \( \psi_{enc} \), which transforms \( S \) into video-level features \mbox{\( M \in \mathbb{R}^{T \times D} \)}. Specifically, we use the mBART encoder \cite{mbart}, a 12-layer model initialized with pre-trained parameters from a large corpus. To adapt it to the sign language data while retaining the pre-trained knowledge, we apply LoRA \cite{loralowrankadaptationlarge}, a parameter-efficient fine-tuning method that introduces low-rank matrices into the weights of the model.

\subsection{Pretraining and Alignment}
\textbf{Pseudo-Gloss Pretraining.}
Our goal is to train the image and temporal encoders to learn visual-linguistic representations and ensure that the segment-level features capture the gloss-level information.
We extract pseudo-glosses from target outputs using Parts-of-Speech (POS) tagging, keeping key word categories [\texttt{Noun}, \texttt{Num}, \texttt{Adv}, \texttt{Pron}, \texttt{Propn}, \texttt{Adj}, \texttt{Verb}]. The lemmatized German words that meet this filter criterion are treated as pseudo-glosses. Each pseudo-gloss is represented by a 300-dimensional fastText~\cite{fasttext} embedding, considered as prototypes, with an additional zero vector for non-sign components, forming the prototype matrix \mbox{\( P \in \mathbb{R}^{D^{'} \times U} \)}, where $D^{'}=300$ and $U$ are the embedding size and number of all pseudo-glosses exist in the dataset.  
As illustrated in Fig. \ref{fig:example}, segment-level features \mbox{\( Z = \{z_{0}, z_{1},...,z_{T} \} \)} are projected to \mbox{\( Z' \in \mathbb{R}^{T \times D^{'}} \)} via a linear layer. We compute the cosine similarity between each projected feature and the pseudo-gloss representations:  
\begin{equation}
    s_{i} = \texttt{sim}(z'_{i}, P) = \frac{z'_{i} \cdot P}{\|z'_{i}\| \|P\|}
\end{equation}
yielding similarity scores \mbox{\( S \in \mathbb{R}^{T \times U} \)}, where higher values indicate stronger alignment.  

To refine prototype localization, we apply temperature-scaled \texttt{softmax} operations along the temporal and prototype axes:  
\begin{equation}
    \hat{S}^{T} = \texttt{softmax} \left(S/\tau_{T}\right), \quad \hat{S}^{U} = \texttt{softmax} \left(S/\tau_{U}\right)
\end{equation}
where learnable scaling factors \( \tau_T \) and \( \tau_U \) control their influence. Element-wise multiplication of \( \hat{S}^{T} \) and \( \hat{S}^{U} \) forms \mbox{\( E \in \mathbb{R}^{T \times U} \)}, which is aggregated over time to produce \mbox{\( \hat{E} \in \mathbb{R}^{U} \)}:  
\begin{equation}
    \hat{E}_{j} = \sum_{i=0}^{T} E_{i,j}
\end{equation}
where high \( \hat{E}_{j} \) values indicate the presence of a pseudo-gloss. We optimize this using binary cross-entropy loss (BCE), assigning 1 if a pseudo-gloss appears in the sign video and 0 otherwise, resulting in an one-hot encoded label $H_{j} \in \mathbb{R}^{U}$. As shown in Fig. \ref{fig:example}, in the pseudo-gloss pretraining stage, for each input video frames, \mbox{$L_{\text{psp}} = \texttt{BCE}(\hat{E}_{j}, H_{j})$}.

\begin{table*}[!t]
\caption{Experimental results on PHOENIX14T dataset. The best results for gloss-free models are highlighted in bold, while the secondbest results are underlined.}
    \centering
    \resizebox{0.7\linewidth}{!}{
    \begin{tabular}{ lccccc }
    
        \hline
        \multirow{2}{*}{\textbf{Method}} & \multicolumn{5}{c}{\textbf{Test Set}}  \\ \cline{2-6} &\rule{0pt}{1em} \textbf{BLEU-1} & \textbf{BLEU-2} & \textbf{BLEU-3} & \textbf{BLEU-4} & \textbf{ROUGE} \\ \hline

        \rowcolor[HTML]{D2D2D2}
        \multicolumn{6}{c}{\textbf{Gloss-based}} \\ \hline
        MMTLB\cite{mmtlb} &  53.97 & 41.75 & 33.84 & 28.39 & 52.65 \\
        TS-SLT\cite{ts-slt} &  54.90 & 42.43 & 34.46 & 28.95 & 53.48 \\ 
        SLTUNET\cite{sltunet}  & 52.92 & 41.76 & 33.99 & 28.47 & 52.11 \\ 
        \hline

        \rowcolor[HTML]{D2D2D2}
        \multicolumn{6}{c}{\textbf{Weakly supervised gloss-free}} \\ \hline  
        TSPNet\cite{tspnet} & 36.10 & 23.12 & 16.88 & 13.41 & 34.96 \\
        GASLT\cite{gaslt} & 39.07 & 26.74 & 21.86 & 15.74 & 39.86 \\
        VAP\cite{vap} & 53.07 & - & - & 26.16 & 51.28\\
        \hline
        \rowcolor[HTML]{D2D2D2}
        \multicolumn{6}{c}{\textbf{Gloss-free}} \\ \hline
        NSLT\cite{nslt} &27.10 &15.61 &10.82 &8.35 &29.70\\
        NSLT+Luong\cite{luong} & 29.86 & 17.52 & 11.96 & 9.00 & 30.70 \\
        CSGCR\cite{csgcr} & 36.71 & 25.40 & 18.86 & 15.18 & 38.85 \\
        GFSLT-VLP\cite{gfslt-vlp} & 43.71 & 33.18 & 26.11 & 21.44 & 42.49 \\
        Sign2GPT(w/PGP)\cite{sign2gpt} & \underline{49.54} & \underline{35.96} & \underline{28.83} & 22.52 & \underline{48.90} \\
        FLa-LLM\cite{fla-llm} & 46.29 & 35.33 & 28.03 & \underline{23.09} & 45.27 \\
        
        \cdashline{1-6}
        \textbf{Ours} &\rule{0pt}{1em}\textbf{49.78} & \textbf{36.07} & \textbf{29.50} & \textbf{23.15} & \textbf{49.10} \\
        \hline
    \end{tabular}}
    \label{tab:sota}
\end{table*}

\textbf{Video-Language Alignment.} Segment-level features \( S \) capture visual information aligned with gloss-level representations and are transformed into video-level features via an LLM encoder. To bridge the video-text modality gap, we employ video-language alignment. 
Inspired by CLIP \cite{clip}, we use contrastive learning to bring video features closer to the corresponding representations of target spoken sentences and move them away from unrelated representations.
Given a target spoken sentence \( TS_{j} \) for video features \( M_{j} \), we embed it into text features \mbox{\( L_{j} \in \mathbb{R}^{\overline{T} \times D} \)} using a frozen 12-layer mBART encoder \( \psi_{te} \) \cite{mbart}, where $\overline{T}$ and $D$ are the number of tokens and dimension of embeddings, respectively. To align video and text, we apply average pooling over frame and token sequences, producing global video \mbox{\( \tilde{M}_{j} \in \mathbb{R}^{D} \)} and sentence \mbox{\( \tilde{L}_{j} \in \mathbb{R}^{D} \)} features, as demonstrated in Fig. \ref{fig:example}. Contrastive learning is then applied:  
\begin{equation}
    \begin{split}
        \mathcal{L}_{\text{align}} = -&\frac{1}{2B}\Big(\sum^{B}_{j=1}log\frac{\exp(\texttt{sim}(\tilde{M}_{j}, \tilde{L}_{j})/\tau)}{\sum^{B}_{k=1}\exp(\texttt{sim}(\tilde{M}_{j}, \tilde{L}_{k})/\tau)} \\
        +& \sum^{B}_{j=1}log\frac{\exp(\texttt{sim}(\tilde{L}_{j}, \tilde{M}_{j})/\tau)}{\sum^{B}_{k=1}\exp(\texttt{sim}(\tilde{L}_{j}, \tilde{M}_{k})/\tau)}\Big)
    \end{split}
\end{equation}

where \( \texttt{sim}(x, y) \) is the cosine similarity, and \( \tau \) is a learnable temperature parameter.  
The final pre-training loss is:  
\begin{equation}
\mathcal{L}_{\text{pre-train}} = \mathcal{L}_{\text{align}} + \lambda\mathcal{L}_{\text{psp}}    \label{eq:loss}
\end{equation}

where \( \lambda \) controls the balance between alignment and prototype supervision losses.

\subsection{Sign Language Translation}
For end-to-end gloss-free SLT, we initialize with pre-trained networks and introduce a linear mapper after the temporal encoder for better adaptation. Given a sign language video \( V_i \), we extract video-level features \( M_i \), which are then processed by a 12-layer mBART decoder \cite{mbart} \( \psi_{dec} \) to generate the spoken sentence \mbox{\( \widehat{TS}_i = (\widehat{TS}_{i,1}, \dots, \widehat{TS}_{i,T}) \)}.  

\( \psi_{dec} \) follows an autoregressive approach, starting with \texttt{<bos>} and generating tokens sequentially until \texttt{<eos>}. Training minimizes cross-entropy loss between predicted tokens \( \widehat{TS}_{i,j} \) and ground truth \( TS_{i,j} \):  
\begin{equation}
    \mathcal{L}_{SLT_{i}} = - \sum^{T}_{j=1}log\hspace{2pt} p(\widehat{TS}_{i,j}|TS_{i,1:j-1}, V_{i})
\end{equation}
\section{Experiments and Results}
\subsection{Dataset and Metrics}

\textbf{Dataset.} 
Following previous studies, we evaluate our method with the PHOENIX14T SLT benchmark \cite{nslt}, a dataset in German sign language with three years of news and weather forecasts. It includes 7,096 training, 519 development and 642 test videos.

\textbf{Evaluation Metrics.} We evaluate SLT using BLEU \cite{bleu} and ROUGE \cite{rouge} metrics. \mbox{BLEU-n} assesses \mbox{n-gram} precision using the geometric mean, while ROUGE calculates the F1 score based on the longest common subsequence, taking into account word order and position.

\subsection{Model and Implementation details}
\textbf{Model.} 
To address memory constraints, we apply LoRA adapters to the top three layers of the visual encoder. 
The temporal encoder is a four-layer transformer with a hidden dimension of 512, 8 attention heads and an intermediate size of 2048, with temporal downsampling applied after the second layer. LoRA \cite{loralowrankadaptationlarge} is applied to the LLM encoder and decoder modules, targeting $q_{\text{proj}}$ and $v_{\text{proj}}$ layers with a rank of 16, a scaling factor (LoRA alpha) of 32 and a dropout rate of 0.1. Training is optimized with the XFormers \cite{xFormers2022} library with Flash Attention~\cite{dao2023flashattention2fasterattentionbetter} for improved efficiency.

\textbf{Pre-training.} We conduct respective pretraining tasks on the training sets of the \mbox{PHOENIX14T} datasets. The number of total pseudo-glosses is $2,322$ and we initialize the prototype ($\tau_{U}$) and time temperature ($\tau_{T}$) to 0.1. Pre-training is conducted for 100 epochs, and the model weights corresponding to the lowest total validation loss (\(\mathcal{L_{\text{pre-train}}}\)) are saved for fine-tuning.

\textbf{Fine-tuning.} 
As mentioned before, we introduce a mapper, a single linear layer, to improve the adaptation to the translation task. According to Cambrian-1 \cite{cambrian1fullyopenvisioncentric}, a two-stage fine-tuning proves beneficial. In the first stage, all modules except the mapper and the decoder are frozen and training is performed for 100 epochs. In the second stage, we fine-tune all trainable parameters in our proposed method for another 200 epochs.
The model is pre-trained and fine-tuned with a batch size of 16 on one A100 GPU. The input sequences are first resized into \mbox{$256 \times 256$}, and then randomly/centrally cropped into \mbox{$224 \times 224$} during training/inference.   
We use the \textit{AdamW} \cite{adamw} optimizer with a learning rate of \mbox{$3 \times 10^{-4}$}
and weight decay of $0.001$. To enhance regularization, we apply gradient norm clipping with a threshold of 1.0 and use a one-cycle cosine learning rate scheduler with a five-epoch warmup. In addition, data augmentation techniques, including random rotation, resizing, and translation, are applied to all frames within the video sequences with a $50\%$ probability. 

\subsection{Comparison}
\textbf{Results on PHOENIX14T.} Tab. \ref{tab:sota} compares gloss-based, weakly supervised gloss-free, and gloss-free methods on the PHOENIX14T test dataset. Our method outperforms other gloss-free SLT models in terms of BLEU-4 and achieves competitive results in all metrics. In particular, it outperforms GFSLT-VLP~\cite{gfslt-vlp} and Sign2GPT~\cite{sign2gpt} in BLEU-4 score on the test set by +1.71 and +0.63, respectively. Moreover, the improvement in \mbox{BLEU-4} and ROUGE score highlights the ability of our model to effectively capture long phrases and contextual information. As shown in Tab.~\ref{tab:sota}, our performance still lags behind VAP\cite{vap}, a weakly supervised method, because it simplifies the translation task by using an intermediate gloss representation. This provides strong supervisory signals that our end-to-end, gloss-free model does not use, forcing it to learn a direct mapping from raw video to text, which is a significantly more complex challenge.
\begin{table}[]
\caption{Sensitivity analysis of $\lambda$ on the SLT results.}
\centering
\begin{tabular}{c|cccc}
\hline
Param & BLEU-1 & BLEU-2 & BLEU-3 & BLEU-4 \\ \hline
$\lambda=0.0$& 43.71 & 34.23 & 27.2 & 21.82 \\
$\lambda=0.5$& 48.2 & 35.10 &  28.43 & 22.50 \\
$\lambda=1.0$&\textbf{50.28}&\textbf{38.07}& \textbf{30.48}& \textbf{23.68}\\ \hline
\end{tabular}
\label{tab:ablation}
\end{table}
\begin{table}[]
\centering
\caption{Number of parameters during pre-training and fine-tuning}
\resizebox{0.95\linewidth}{!}{
\begin{tabular}{l|cl|cl}
\hline
\multirow{2}{*}{\textbf{Methods}} & \multicolumn{2}{c|}{\textbf{Pre-training}} & \multicolumn{2}{c}{\textbf{Fine-tuning}}\\ \cline{2-5} 
& \multicolumn{1}{c|}{\# \textbf{Trainable}} & \# \textbf{Total}    & \multicolumn{1}{c|}{\# \textbf{Trainable}} & \# \textbf{Total}      \\ \hline
GFSLT-VLP~\cite{gfslt-vlp} &  \multicolumn{1}{l|}{215,597,633}  & 215,597,633 & \multicolumn{1}{l|}{114,835,008}  & 114,835,008   \\
Sign2GPT~\cite{sign2gpt}  & \multicolumn{1}{l|}{13,889,604}   & 35,946,180  & \multicolumn{1}{l|}{16,689,024}   & 1,771,652,608 \\
Ours& \multicolumn{1}{l|}{15,200,325}   & 346,123,973 & \multicolumn{1}{l|}{16,819,712}   & 395,924,608  \\ \hline
\end{tabular}}
\label{tab:param}
\end{table}
\subsection{Ablation Study}
We conducted an experiment on the PHOENIX14T development set to evaluate the impact of the parameter \(\lambda\) in Eq. \ref{eq:loss} on the final results. As shown in Tab. \ref{tab:ablation}, increasing the weight of the pseudo-gloss pre-training loss (\(\mathcal{L_{\text{psp}}}\)) within the total pre-training loss (\(\mathcal{L_{\text{pre-train}}}\)) leads to better performance.
We can see in Tab. \ref{tab:param}, our model achieves superior efficiency with 15.2 million trainable parameters in pre-training (vs. \mbox{GFSLT-VLP’s} 215.6 million) and 395.9 million total parameters in fine-tuning (vs. Sign2GPT’s 1.77 billion total parameters), balancing performance, reduced computational load, and adaptability for real-world use.
\section{CONCLUSIONS AND FUTURE WORKS}
We introduced a hierarchical pretraining strategy for gloss-free SLT, leveraging pseudo-gloss and contrastive video-language alignment to bridge modality gaps. Our method enhances translation performance while maintaining efficiency. Future work can explore further refinements in feature alignment and scalability.

\begin{acks}
    This work was supported by the SNSF project ‘SMILE II’ (CRSII5 193686), the Innosuisse IICT Flagship (PFFS-21-47), EPSRC grant APP24554 (SignGPT-EP/Z535370/1) and through funding
from Google.org via the AI for Global Goals scheme. This work reflects only the author’s views and the funders are not responsible for any use that may be made of the information it contains.
\end{acks}

\bibliographystyle{acm}
\bibliography{main}










\end{document}